\def\year{2019}\relax
\documentclass[letterpaper]{article} 
\usepackage{aaai19}  
\usepackage{times}  
\usepackage{helvet}  
\usepackage{courier}  
\usepackage{url}  
\usepackage{graphicx}  
\usepackage{graphicx}
\usepackage{subcaption}
\usepackage{amsmath}
\usepackage{amsfonts}
\usepackage{algorithm}
\usepackage[noend]{algpseudocode}
\usepackage[capitalise]{cleveref}
\usepackage{booktabs}
\usepackage{multirow}
\usepackage{tikz}
\usetikzlibrary{positioning,calc,matrix,arrows,shapes.arrows}

\usepackage{pifont}
\newcommand{\cmark}{\ding{51}}

\makeatletter
\def\BState{\State\hskip-\ALG@thistlm}
\makeatother

\definecolor{arrowblue}{RGB}{91,155,230}
\definecolor{arrowgreen}{RGB}{112,173,71}

\begin{document}
%
\title{Efficient Image Retrieval via Decoupling Diffusion \\ into Online and Offline Processing}
\author{Fan Yang\textsuperscript{1,2}, Ryota Hinami\textsuperscript{1,2}, Yusuke Matsui\textsuperscript{1}, Steven Ly\textsuperscript{2,3}, Shin'ichi Satoh\textsuperscript{2,1} \\
\textsuperscript{1}The University of Tokyo, Japan \\
\textsuperscript{2} National Institute of Informatics, Japan \\
\textsuperscript{3} University of Southern California, USA
}

\maketitle
\begin{abstract}
Diffusion is commonly used as a ranking or re-ranking method in retrieval tasks to achieve higher retrieval performance, and has attracted lots of attention in recent years.
A downside to diffusion is that it performs slowly in comparison to the naive $k$-NN search, which causes a non-trivial online computational cost on large datasets.
To overcome this weakness, we propose a novel diffusion technique in this paper.
In our work, instead of applying diffusion to the query, we pre-compute the diffusion results of each element in the database, making the online search a simple linear combination on top of the $k$-NN search process.
Our proposed method becomes 10$\sim$ times faster in terms of online search speed.
Moreover, we propose to use late truncation instead of early truncation in previous works to achieve better retrieval performance.
\end{abstract}

\section{Introduction}

The success of deep neural networks on feature representation has led it to become a standard technique in image retrieval.
Models pre-trained on popular datasets such as ImageNet~\cite{deng2009imagenet}, Landmarks~\cite{babenko2014neural} etc. can be used to extract features of images.
Particularly, convolutional layers have been proved to be most beneficial at retrieving images~\cite{babenko2014neural,radenovic2016cnn,gordo2016deep,razavian2016visual}.
Nearest neighbor search is then used on the feature vectors to find the most similar images to a query.

Datasets are usually scraped from the internet, resulting in images with the same object/landmark shown in a variety of angles, lighting, and other conditions.
The diversity often results in a manifolds in the feature space that are not conducive to ranking using a distance-based metric.
Unlike the rigid distance metric used in $k$-NN search, diffusion~\cite{zhou2004ranking,zhou2004learning,donoser2013diffusion,grady2006random} exploits the intrinsic manifold structure of data based on a neighborhood graph.
Such a graph consists of nodes and edges, where each node represent a feature vector from the database, with the edges connect each node to its neighbors with corresponding weights proportional to the pairwise affinities between nodes.
Using this graph, diffusion performs a restartable random walk given a query as the initial state.
The final state of random walk can be viewed as ranking scores showing the similarities of each image in database to the query.
To obtain the convergence of the final state, there are two main approaches: running iterative random walk or computing the convergence state by the closed-form theorem proposed in~\cite{zhou2004ranking}.
Diffusion has demonstrated its potential in improving retrieval performance~\cite{donoser2013diffusion,iscen2017efficient,radenovic2018revisiting}, and is also utilized in other fields such as unsupervised learning~\cite{iscen2018mining}.
Recently, other works have attempted to improve the efficiency of diffusion~\cite{iscen2017efficient}, but their speed up is still not sufficient enough to handle the amount of queries found in large-scale image retrieval datasets.

\begin{figure}[t!]
  \centering
  \includegraphics[width=.8\linewidth]{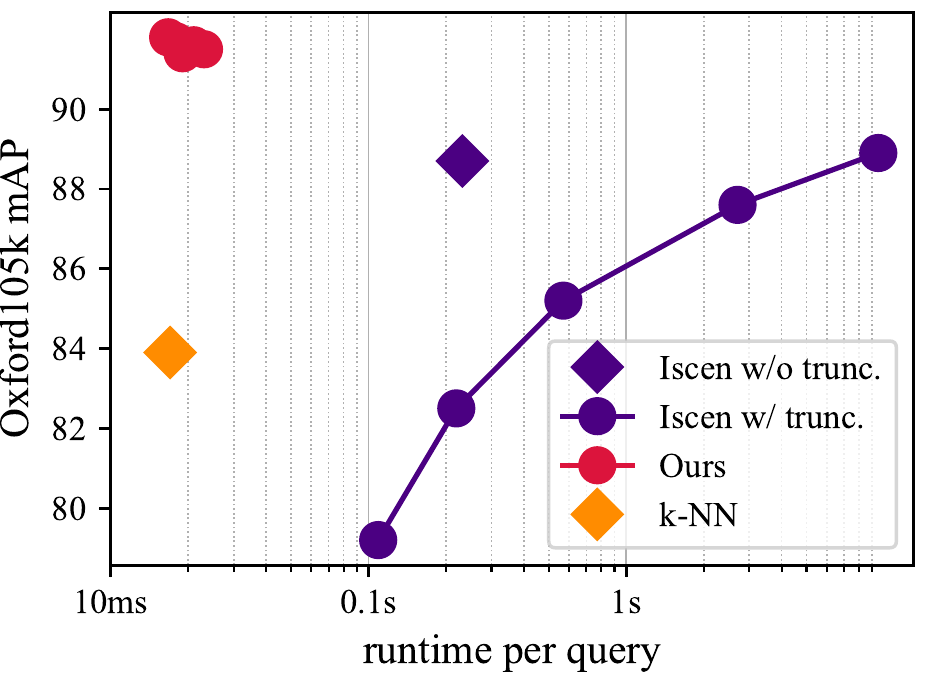}
  \caption{The efficiency vs. performance comparison between $k$-NN, Iscen's method~\cite{iscen2017efficient}, and our proposed method on Oxford105k dataset with global image features. 
  Our proposed method achieves better retrieval performance than diffusion by Iscen et al. with almost the same search speed ($\sim$20ms per query) as $k$-NN search. We show the results of varying truncation sizes $L$.}
  \label{fig:top}
\vspace{-4mm}
\end{figure}

We notice that the main bottlenecks in speed of online diffusion processes come from the random walk and preparation steps.
Inspired by the closed-form solution of diffusion, we find that the diffusion for each query can be converted to a linear combination of the pre-computed diffusion results of all database elements.
Following this observation, our proposed method completely removes the random walk from the online stage.
As a result, our work is able to improve the efficiency of the diffusion process by a factor of ten in a large-scale image retrieval setting.

In addition, the previous versions of diffusion utilize an early truncation that happens before the affinity matrix normalization process.
In our proposed process, we propose to perform a late truncation after normalization, that results in significantly better performance.
\cref{fig:top} summarizes the efficiency and performance of the aforementioned methods.
The source code to replicate our experiments is available at \url{https://github.com/fyang93/diffusion}.

\section{Related Works}

Although originally developed for ranking on manifolds~\cite{page1999pagerank,zhou2004ranking,donoser2013diffusion}, diffusion was soon applied to classification~\cite{zhou2004learning}, and image segmentation~\cite{grady2006random}.
Recently, some variants of diffusion~\cite{bai2017regularized,bai2017ensemble,bai2018regularized,bai2019automatic} are also developed to conduct diffusion processes on tensor product graph with a smoothness criterion, showing its potential for image retrieval tasks as a re-ranking method.

Query expansion, a common technique in image retrieval, can improve retrieval performance during query time.
Average query expansion (AQE)~\cite{chum2007total,iscen2017efficient}, a popular type of query expansion because of its simplicity, averages the features of the query's nearest neighbors to form a new query to run search again. 
When AQE is applied iteratively, the recomputation of the query is akin to traveling along the manifolds of the feature space.
Although this traversal is similar to diffusion, AQE only utilizes the relationships between query and database images, but not between each of the database images with each other.
With prior knowledge of the relationships between all of the database images, diffusion is thus better able to exploit the manifolds in the feature space than query expansion can. 

In previous works of diffusion, the query is provided as a part of the database.
However, in a real-world setting, queries are unavailable until they are issued by users.
To tackle this issue without introducing any computational overhead, \cite{iscen2017efficient} uses the short list of $k$-NN search results to form a sparse initial state vector, instead of using a one-hot vector as the initial state.
As a consequence, queries are not included in the neighborhood graph.
The downside to this is that the graph needs to be stored and loaded during the search stage for random walk, which is both memory and computationally inefficient.
Since the previous methods were evaluated on the Oxford~\cite{philbin2007object} and Paris~\cite{philbin2008lost} datasets, smaller datasets only containing 55 queries, the inefficiency of those methods did not have much impact on the total computation time.
When these methods are used on large-scale datasets with many queries, the inefficiency during online search becomes magnified and intractable.

To tackle this inefficiency, past efforts have been made to scale diffusion up to handle larger datasets.
~\cite{dong2011efficient} proposed to accelerate the construction of the affinity matrix denoting the graph.
Iscen et al. reported that Dong's method is orders of magnitude faster than exhaustive search with only limited decreases in performance~\cite{iscen2017efficient}.
Another approach to improve efficency is using approximate nearest neighbor search (ANN).
Compared to constructing the graph by exhaustive $k$-NN search, ANN search is faster and provides comparable accuracy~\cite{jegou2011product,ge2014optimized}.
Most recently, \cite{iscen2018fast} approximated the affinity matrix by using a low-rank spectral decomposition to reduce the online computational cost.
However, this method did not result in much improvement in terms of retrieval performance.

\section{Preliminaries of Diffusion}
\label{sec:preliminaries}

There are two main approaches to conducting diffusion: through iterative updates or solving the closed form directly.
Both Zhou et al. and Donoser et al. describe diffusion as a mechanism for spreading the query similarities over the manifolds~\cite{zhou2004ranking,donoser2013diffusion},
while Iscen et al. utilizes the closed form theorem in~\cite{zhou2004ranking} and proposed an efficient solution~\cite{iscen2017efficient}.
We mainly follow the steps from \cite{zhou2004learning} and ~\cite{iscen2017efficient} below.

\subsubsection{Problem setup.}

For image retrieval tasks, we define a database as $\chi=\{\mathbf{x}_1,\dots,\mathbf{x}_n\}\subset\mathbb{R}^d$, where each $\mathbf{x}_i$ is a feature vector.
Images can be represented by a global feature that corresponds to the entire image, or multiple regional features corresponding to different regions of the image.
In the following equations, $\mathbf{x}_i$ can stand for either of these representations.

For most public datasets in the retrieval field, query and database images are both available.
In our following example, queries are unseen to us until provided by users.
We denote the query as $\mathcal{Q}=\{\mathbf{q}_1,\dots,\mathbf{q}_m\}\subset\mathbb{R}^d$, where $m=1$ when the query is described by a global feature and $m>1$ when it contains regional features.

\subsubsection{Graph construction.}
\label{sec:graph_construction}
For simplicity, we consider an example where we handle only one query image $\mathcal{Q}$ and include it into the database.
The entire set is defined as $\bar{\chi}=\{\mathbf{q}_1,\dots,\mathbf{q}_m,\mathbf{x}_1,\dots,\mathbf{x}_n\}$, and we denote $i$-th element in $\bar{\chi}$ as $\bar{\chi}_i$.
In addition, a local constraint is adopted so that the graph only contains similarities between pairs of elements that are nearest neighbors to each other according to~\cite{iscen2017efficient}.
The affinity matrix is defined as $\mathbf{A}=(a_{ij})\in\mathbb{R}^{(n+m)\times(n+m)}$, where each element is obtained by
\begin{equation}
\label{equ:affinity}
\resizebox{.9\hsize}{!}{$
a_{ij} =
\left\{
\begin{aligned}
    & s(\bar{\chi}_i, \bar{\chi}_j) &i\neq j,\bar{\chi}_i\in\textrm{NN}_k(\bar{\chi}_j), \bar{\chi}_j\in\textrm{NN}_k(\bar{\chi}_i) \\
    & 0 &\textrm{otherwise} \qquad\qquad\qquad\qquad\qquad~~~~
\end{aligned}
\right.,
$}
\end{equation}
$\forall i,j \in \{1,\dots,n+m\}$, denoting $\textrm{NN}_k(\mathbf{x})$ the $k$-NNs of $\mathbf{x}$.
Since the similarity metric $s$ is usually symmetric and positive, $\mathbf{A}$ is a symmetric matrix.
\cref{equ:affinity} allows $\mathbf{A}$ to be sparse, providing memory and computational efficiency.

The degree matrix $\mathbf{D}$ is a diagonal matrix and each diagonal element is the corresponding row-wise sum of $\mathbf{A}$, \emph{i.e.} the element $d_{ii}$ in $\mathbf{D}$ is defined as $\sum_{j=1}^{n+m}a_{ij}$.
It's later used to symmetrically normalize $\mathbf{A}$ into the stochastic matrix $\mathbf{S}$:
\begin{equation}
\label{equ:normalization}
  \mathbf{S} = \mathbf{D}^{-1/2} \mathbf{A} \mathbf{D}^{-1/2}.
\end{equation}
$\mathbf{S}$ is a variant of the typical transition matrix $\mathbf{D}^{-1}\mathbf{A}$, and both have the same eigenvalues and eigenvectors~\cite{donoser2013diffusion}.

\subsubsection{Random walk.}
After the graph construction, the random walk is performed until it reaches a convergence state, resulting in final ranking scores for each of the images in the gallery.
For the $t$-th step of random walk, the state is recorded in a vector $\mathbf{f}^t=[{\mathbf{f}_q^t}^\top, {\mathbf{f}_d^t}^\top]^\top \in \mathbb{R}^{n+m}$, where $\mathbf{f}_q^t\in \mathbb{R}^m, \mathbf{f}_d^t\in\mathbb{R}^n$.
We set the initial state to be a $m$-hot vector where $\mathbf{f}_q^0=\mathbf{1}_m, \mathbf{f}_d^0=\mathbf{0}_n$.
The random walk iterates the following step:
\begin{equation}
  \mathbf{f}^{t+1} = \alpha \mathbf{S} \mathbf{f}^t + (1-\alpha) \mathbf{f}^0, \quad \alpha \in (0, 1).
\end{equation}
Essentially, there is a probability $\alpha$ to randomly walk from the current state $\mathbf{f}^t$ or $1-\alpha$ to restart from the initial state $\mathbf{f}^0$.
Given the fact that $\alpha \in (0,1)$ and the abstract eigenvalues of $\mathbf{S}$ is no larger than 1 according to the Perron-Frobenius theorem, this iteration converges to a closed-form solution~\cite{zhou2004ranking}:
\begin{equation}
  \mathbf{f}^* = (1-\alpha)(\mathbf{I}-\alpha \mathbf{S})^{-1}\mathbf{f}^0.
\end{equation}
After convergence, the values in $\mathbf{f}^*$ contain the similarities of each database element to the query, which will be used as ranking scores for re-ranking.

\subsubsection{Decomposition.}

The above steps incorporate the query into the graph during the diffusion process.
Grady proposed to decompose queries from the above operations~\cite{grady2006random}, and his technique was recently followed by~\cite{iscen2017efficient}.

Note, the closed-form solution $\mathbf{f}^*\in\mathbb{R}^{n+m}$ contains the ranking scores on both the query and database elements,
but for the task of image retrieval, we only care about the ranking scores for database elements.
This leads to the decomposition of the query and database ranking scores,
so that the matrix $\mathbf{S}$ is split into 4 blocks
\begin{equation}
  \mathbf{S} = \begin{bmatrix}
      \mathbf{S}_{qq} & \mathbf{S}_{qd} \\
      \mathbf{S}_{dq} & \mathbf{S}_{dd}
  \end{bmatrix},
\end{equation}
where $\mathbf{S}_{qq}\in\mathbb{R}^{m\times m},\mathbf{S}_{qd}\in\mathbb{R}^{m\times n},\mathbf{S}_{dq}\in\mathbb{R}^{n\times m}$, and $\mathbf{S}_{dd}\in\mathbb{R}^{n\times n}$.
The decomposed solution then becomes
\begin{equation}
\mathbf{f}_d^*=(1-\alpha)(\mathbf{I}-\alpha\mathbf{S}_{dd})^{-1}\mathbf{S}_{dq}\mathbf{f}_q^0 \in \mathbb{R}^n,
\end{equation}
where $\mathbf{S}_{dd}$ can be viewed as the transition matrix for random walk on the database side, and $\mathbf{S}_{dq}=\mathbf{S}_{qd}^\top$ consists of normalized similarities between the query and its nearest neighbors.
Subsequently, we can then obtain the cleaner form:
\begin{equation}
\label{equ:clean_form}
\mathbf{f}_d^* \propto \mathcal{L}_\alpha^{-1} \mathbf{y},
\end{equation}
where $\mathcal{L}_\alpha = \mathbf{I} - \alpha \mathbf{S}_{dd}\in\mathbb{R}^{n\times n}, \mathbf{y} = \mathbf{S}_{dq}\mathbf{f}_q^0\in\mathbb{R}^n$.
and we can ignore the constant $1-\alpha$ since scores are used for ranking.
\subsubsection{Truncation.}
An optional truncation step is used on large datasets, where the dataset is truncated to a smaller subset before normalization and random walk.
Iscen et al. conduct truncation with only global features representing entire images~\cite{iscen2017efficient}.
Given the global feature vector $\mathbf{q}$ of a new query ($m=1, \mathbf{q}=\mathbf{q}_1$), the indexes $\mathcal{I}=\textrm{NN}_L^{\textrm{ID}}(\mathbf{q})$ of features corresponding to the top ranked images is retrieved by $L$-NN search, where $L$ is a limiting constant that defines the maximum size of the subgraph (truncated graph).
The affinity matrix denoting the subgraph is defined as $\hat{\mathbf{A}}\in\mathbb{R}^{L\times L}$, and each element $\hat{a}_{ij}$ in $\hat{\mathbf{A}}$ satisfies
\begin{equation}
\resizebox{.9\hsize}{!}{$
\hat{a}_{ij} =
\left\{
\begin{aligned}
  & s(\chi_{\mathcal{I}_i}, \chi_{\mathcal{I}_j}) & \mathcal{I}_i \neq \mathcal{I}_j, \chi_{\mathcal{I}_i} \in \textrm{NN}_k(\chi_{\mathcal{I}_j}), \chi_{\mathcal{I}_j} \in \textrm{NN}_k(\chi_{\mathcal{I}_i}) \\
  & 0 & \textrm{otherwise} \qquad\qquad\qquad\qquad\qquad\qquad~~~~~~~~
\end{aligned}
\right. ,
$}
\end{equation}
$\forall i,j \in \{1,\dots,L\}$, where $\mathcal{I}_i$ is the $i$-th index in the set $\mathcal{I}$.
After truncation, $\hat{\mathbf{A}}$ is normalized into a stochastic matrix $\hat{\mathbf{S}}$ and then random walk is subsequently performed on $\hat{\mathbf{S}}$.
We refer to the process of normalizing the truncated graph as subgraph normalization throughout the rest of the paper.

\begin{figure}[t]
\centering
\begin{tikzpicture}[>=latex]
\tikzset{
ba/.style={single arrow, draw=arrowblue, fill=arrowblue, text=white, font=\small, ultra thick, text width=1.5cm, align=center},
ga/.style={single arrow, draw=arrowgreen, fill=arrowgreen, text=white, font=\small, ultra thick, text width=1.5cm, align=center},
}

\node (A) {$\mathbf{A}$};
\node[right=2.5cm of A] (S) {$\mathbf{S}$};
\node[right=2.5cm of S] (L) {$\mathcal{L}_\alpha$};
\node[above=1.8cm of A] (At) {$\hat{\mathbf{A}}$};
\node[above=1.8cm of S] (St) {$\hat{\mathbf{S}}$};
\node[above=1.8cm of L] (Lt) {$\hat{\mathcal{L}}_\alpha$};

\node[ga, right=0.3cm of A] {normalize};
\node[ga, right=0.1cm of S] {create};
\node[ga, above=0.8cm of L, anchor=center, rotate=90, text width=1cm] {truncate};
\node[ba, right=0.3cm of At] {normalize};
\node[ba, right=0.1cm of St] {create};
\node[ba, above=0.8cm of A, anchor=center, rotate=90, text width=1cm] {truncate};

\end{tikzpicture}
\caption{Comparison between our implementation (green) and previous works' (blue). Our method truncates late while previous methods truncate early during diffusion.}
\label{fig:trunc_order}
\vspace{-3mm}
\end{figure}
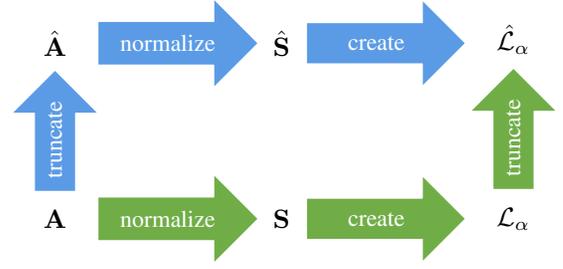

\section{Proposed Method}

We propose a remarkably fast diffusion achieving state-of-the-art retrieval performance.
Iscen et al. reported that $\mathcal{L}_\alpha^{-1}$ is not sparse like $\mathcal{L}_\alpha$ making it less efficient to compute than using $\mathcal{L}_\alpha$ to solve $\mathcal{L}_\alpha \mathbf{f}_d^*\propto \mathbf{y}$ online~\cite{iscen2017efficient}.
Our method makes it possible to pre-compute and maintain a sparsified $\mathcal{L}_\alpha^{-1}$ offline to achieve better efficiency.
Given a new query, its diffusion result can be obtained by linear combination according to~\cref{equ:clean_form}.
As a result, we achieve a substantial improvement in the online search speed.
Moreover, we argue that the subgraph normalization that takes place in~\cite{iscen2017efficient} has negative effects on the retrieval performance.
After the offline computation to obtain $\mathcal{L}_\alpha$ from the entire matrix $\mathbf{A}$, we apply slicing to $\mathcal{L}_\alpha$ to fetch the values on the corresponding rows and columns limited in the truncation subset.

In the following sections, we compare the time complexity between Iscen's method and our method to analyze the efficiency gains of our method.

\subsection{Complexity analysis of online diffusion}

In prior works, the entire process of diffusion is performed during runtime when queries are processed.
A combination of global and regional features are also used, but for simplicity, we choose to analyze Iscen's online diffusion with only global features in this section.

Given the global feature $\mathbf{q}$ of a new query, the pipeline of online diffusion can be broken down to the following steps:
\begin{enumerate}
\item \emph{Truncation}: the nearest neighbors $\textrm{NN}_L(\mathbf{q})\subset\chi$ of $\mathbf{q}$ is obtained by $k$-NN search for truncation
\item \emph{Graph construction}: the truncated affinity matrix $\hat{\mathbf{A}}$ denoting subgraph is constructed for the subset $\textrm{NN}_L(\mathbf{q})$ with a reciprocity check,
then \textit{subgraph normalization} is applied to form the matrix $\hat{\mathbf{S}}$ and $\hat{\mathcal{L}}_\alpha$ are created afterward
\item \emph{Initialization}: the vector $\mathbf{y}=\mathbf{S}_{dq}\mathbf{f}_q^0$ contains the similarities between the query and its nearest neighbors, which is subsequently truncated to fit the size of $\hat{\mathcal{L}}_\alpha$
\item \textit{Random walk}: the convergence state is solved by~\cref{equ:clean_form} to obtain the result of diffusion by conjugate gradient
\end{enumerate}
The above steps include $k$-NN search, which is responsible for most of the time needed to process those steps.
Recent $k$-NN search strategies are sufficiently efficient~\cite{jegou2011product,malkov2016efficient,johnson2017billion}, so we do not discuss the complexity which is beyond the scope of this work.
Constructing the affinity matrix denoting the subgraph also requires $k$-NN search,
and the subgraph normalization according to~\cref{equ:normalization} costs $\mathcal{O}(Lk)$ time.
Here, $Lk$ is the number of non-zero entries in the truncated affinity matrix, and $k$ is the parameter for the number of nearest neighbors in $k$-NN search.
In the random walk step, the result is approximated by the early-terminated conjugate gradient (CG)~\cite{nocedal2006numerical,iscen2017efficient}.
Suppose we iterate CG for $\tau$ steps, its time complexity is $\mathcal{O}(Lk\tau)$.

From the above analysis, we can see that the cost to process each query is non-trivial and the most time-consuming steps are the subgraph construction and random walk.
This motivates us to solve these inefficiencies.

\subsection{From online to offline}
We propose a new form of diffusion that moves the \textit{online} steps, processing the heavy computation steps during runtime, to \textit{offline}, pre-computing those steps beforehand.

The right side of~\cref{equ:clean_form} can be considered as a linear combination of column vectors in $\mathcal{L}_\alpha^{-1}$ with weights in vector $\mathbf{y}$.
In other words, the results of any new query is merely the linear combination of the columns of $\mathcal{L}_\alpha^{-1}$ with corresponding weights in $\mathbf{y}$.
Unfortunately, the inverse of a large sparse matrix is hard to compute even though $\mathcal{L}_\alpha$ is positive-definite.
Despite this difficulty, it is still possible to compute the approximate inverse by either global iteration or a column-oriented algorithm, two approaches summarized in~\cite{saad2003iterative}.
Global iteration computes the inverse on the entirety of the matrix, whereas the column-oriented algorithm computes it one column at a time.
Between the two, the column-oriented algorithm is more appealing for parallelism since it computes each column separately.
It also allows us to apply truncation in computing each column to make a sparser structure.
Therefore, we choose to adopt the column-oriented strategy in our proposed method.

To compute each column of $\mathcal{L}_\alpha^{-1}$, we first define a set of vectors $\{\mathbf{b}_1,\dots,\mathbf{b}_n\}$ to be the column vectors of an identity matrix $\mathbf{I}\in\mathbb{R}^{n\times n}$.
Then, according to the closed-form solution in \cref{equ:clean_form} we solve:
\begin{equation}
\label{equ:offline_diffusion}
\mathcal{L}_\alpha \mathbf{c}_i = \mathbf{b}_i,
\end{equation}
 with conjugate gradient (CG)~\cite{nocedal2006numerical}, we obtain $\mathbf{c}_i$, the approximate $i$-th column vector in $\mathcal{L}_\alpha^{-1}$.
Essentially, $\mathbf{c}_i$ can be viewed as the diffusion result of $i$-th database element $\mathbf{b}_i$.
After the database-side diffusion, given a new query, the pipeline becomes
\begin{enumerate}
\item \textit{Initialization}: prepare the initial state vector $\mathbf{y}$ for the query, where the indexes of non-zero entries are $\{i_1, \dots, i_h\}$, and their values are $\{v_1, \dots, v_h\}$
\item \textit{Linear combination}: combine the corresponding columns $\{\mathbf{c}_{i_1}, \dots, \mathbf{c}_{i_h}\}$ from $\mathcal{L}_\alpha^{-1}$ with the values in step 1 to obtain the diffusion result: $\mathbf{f}_d^*\propto\sum_j{v_j \mathbf{c}_{i_j}}$
\end{enumerate}

\begin{figure}[t]
\centering
\begin{tikzpicture}[>=latex]
\tikzstyle{every node}=[font=\small];
\tikzset{
squared/.style={rectangle, draw=blue!60, fill=blue!5, very thick, minimum size=2.5mm},
rounded/.style={rectangle, draw=red!60, fill=red!5, very thick, minimum size=3mm},
ge/.style={minimum size=2mm},
sp/.style={rectangle, draw=red!50, thick, minimum size=2mm},
}

\node[squared] (id1) {$\textrm{NN}^{\textrm{ID}}_L(\mathbf{x}_1)$};
\node[squared, right=5pt of id1] (id2) {$\textrm{NN}^{\textrm{ID}}_L(\mathbf{x}_2)$};
\node[right=5pt of id2] (id3) {...};
\node[squared, right=5pt of id3] (id4) {$\textrm{NN}^{\textrm{ID}}_L(\mathbf{x}_n)$};
\draw[<->, thick] ($(id1.north)+(-18pt,5pt)$) -- node[above]{$L$} ($(id1.north)+(18pt,5pt)$);
\node[rounded, below=10pt of id1] (v1) {$\hat{\mathbf{c}}_1$};
\node[rounded, below=10pt of id2] (v2) {$\hat{\mathbf{c}}_2$};
\node[below=20pt of id3] (v3) {...};
\node[rounded, below=10pt of id4] (v4) {$\hat{\mathbf{c}}_n$};
\draw[-,thick] (id1.south) -- (v1.north);
\draw[-,thick] (id2.south) -- (v2.north);
\draw[-,thick] (id4.south) -- (v4.north);
\draw[<->, very thick] ($(v1.south)+(0,-5pt)$) -- node[below](n){$n$} ($(v4.south)+(0,-5pt)$);

\matrix[below=5pt of n, font=\tiny] (L) [matrix of math nodes, nodes = {ge}, left delimiter  = {[}, right delimiter = {]}]
{
\node[sp] (e1) {\hat{c}_{11}}; & 0 & \ldots & 0  \\
0 & \node[sp] {\hat{c}_{12}}; & \ldots & 0 \\
\node (e2) {\vdots}; & \vdots & \ddots & \vdots  \\
\node[sp] (e3) {\hat{c}_{n1}}; & 0 & \ldots & \node[sp] {\hat{c}_{nn}};  \\
};

\draw[->, thick, color=blue!50] (id1.west) -- ++(-8pt,0) |- ($(e1.west)+(-38pt,0)$) node[above,color=black](first){ID:1} -- (e1.west);
\draw[->, thick, color=blue!50] (id1.west) -- ++(-8pt,0) |- ($(e2.west)+(-43pt,0)$) node[above,color=black]{\vdots} -- ($(e2.west)+(-5pt,0)$);
\draw[->, thick, color=blue!50] (id1.west) -- ++(-8pt,0) |- ($(e3.west)+(-38pt,0)$) node[above,color=black](last){ID:n} -- (e3.west);

\draw[<->, very thick] ($(first.west)+(-10pt,-5pt)$) -- node[sloped,above,color=black]{$L$} ($(last.west)+(-10pt,-5pt)$);

\draw[->, thick, color=red!50] (v1.south) |- (e1.west);
\draw[->, thick, color=red!50] (v1.south) |- ($(e2.west)+(-5pt,0)$);
\draw[->, thick, color=red!50] (v1.south) |- (e3.west);

\node at ($(L.south east)+(18pt,5pt)$) {$\mathcal{L}_\alpha^{-1}$};

\end{tikzpicture}
\caption{The data structure of sparsified matrix $\mathcal{L}_\alpha^{-1}$, where the values in $i$-th column compose the column vector $\hat{\mathbf{c}}_i$ and $\textrm{NN}^{\textrm{ID}}_L(\mathbf{x}_i)$ records the row indexes of each value in $\hat{\mathbf{c}}_i$.
}
\label{fig:lap_inverse}
\vspace{-3mm}
\end{figure}
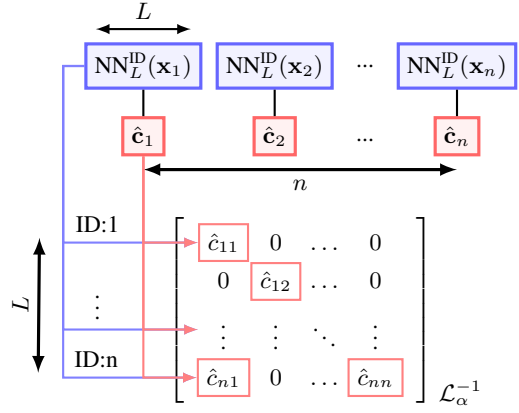

Note that the inverse matrix $\mathcal{L}_\alpha^{-1}$ obtained from the above procedure is a dense matrix, which is less memory efficient.
We propose the sparsified version~(\cref{fig:lap_inverse}) of inverse matrix $\mathcal{L}_\alpha^{-1}$ to provide better memory efficiency in the next section.

\subsection{Database-side truncation}

As we discussed, truncation is crucial for scaling up to large datasets.
The time complexity of diffusion is deeply related to $L$, the size of subgraph after truncation.
Thus, the process of random walk can be accelerated if the database is truncated to a smaller size.
Despite increases in speed, Iscen et al. observed that truncation has a negative effect on the retrieval performance~\cite{iscen2017efficient}.


Contrary to Iscen's findings, we find that truncation by itself is not a detrimental practice.
Rather, the order in which it is applied in relation to normalization is the important factor.
We find that applying normalization after truncation is the reason for the decrease in retrieval performance.
The subgraph after truncation contains incomplete manifolds and the later normalization raises the probabilities to transition to the nodes on such manifolds.
This causes the random walk to be more likely to visit misleading nodes.

To tackle this issue, we normalize the complete matrix $\mathbf{A}$ to build $\mathcal{L}_\alpha$, and then apply late truncation (slicing) to $\mathcal{L}_\alpha$ directly, as shown in~\cref{fig:trunc_order}.
Denoting the indexes of the nearest neighbors of a query as $\mathcal{I}=\textrm{NN}^{\textrm{ID}}_L(\mathbf{q})$, the truncation is applied by
\begin{equation}
\label{equ:truncation}
\hat{l}_{ij} = l_{\mathcal{I}_i \mathcal{I}_j}, \quad \forall i,j \in \{1,\dots,L\},
\end{equation}
where $l_{ij}$ is the element in $\mathcal{L}_\alpha$ while $\hat{l}_{ij}$ stands for the element in $\hat{\mathcal{L}}_\alpha$ after truncation.
From our experiments, such a truncation can provide a significant performance boost.


In previous works, diffusion cannot be performed without the query because truncation process is applied under the guide of queries with a subsequent random walk.
This forces diffusion in those works to be computed online.
Our proposed method differs by instead using the elements in the database themselves as queries, thus moving the entire diffusion process offline.
In addition, truncation is applied to the database elements as previous works applied it on queries.

As an example, we take a database element $\mathbf{x}_i$ and conduct $L$-NN search in $\chi$ to get the indexes $\mathcal{J}=\textrm{NN}_L^{\textrm{ID}}(\mathbf{x}_i)$ of the short list.
Since $\mathcal{L}_\alpha$ can be computed and cached beforehand, we slice $\mathcal{L}_\alpha$ with $\mathcal{J}$ to avoid subgraph normalization.
The one-hot vector $\mathbf{b}_i$ is also sliced to $\hat{\mathbf{b}}_i$ with the indexes $\mathcal{J}$.
Usually the indexes $\textrm{NN}^{\textrm{ID}}_L(\mathbf{x}_i)$ are sorted by the similarities in a descending order.
Thus, the index $i$ of $\mathbf{x}_i$ is always at the top of $\textrm{NN}^{\textrm{ID}}_L(\mathbf{x}_i)$ since it is always most similar to itself.
As a result, the truncated one-hot inital state vector $\hat{\mathbf{b}}_i=[1,0,0,\dots]^\top\in\{0,1\}^L$ is fixed regardless of $i$.
Therefore, we use the same initial state vector for all database-side random walk.
After the truncation on $\mathcal{L}_\alpha$ and $\mathbf{b}_i$, we obtain the truncated $i$-th column vector in $\mathcal{L}_\alpha^{-1}$ by $\hat{\mathcal{L}}_\alpha \hat{\mathbf{c}}_i=\hat{\mathbf{b}}_i$, where $\hat{\mathbf{c}}_i\in\mathbb{R}^L$.
This sparsifies the matrix $\mathcal{L}_\alpha^{-1}$.
\cref{fig:lap_inverse} shows the structure of the sparsified $\mathcal{L}_\alpha^{-1}$.
It costs $\mathcal{O}(Ln)$ memory to store the pre-computed $\mathcal{L}_\alpha^{-1}$, which is more than $\mathcal{O}(L^2)$, the memory usage of Iscen's method.

To summarize, the late truncation directly applied on $\mathcal{L}_\alpha$ removes the negative effects of early truncation, and is completely pre-computed offline.
In addition, the truncated initial state vector is fixed, meaning there is no extra overhead cost to build it.

\begin{algorithm}[t]
\caption{Online search}
\label{alg:online}
\begin{algorithmic}[1]
\State $\textbf{Input } \mathcal{Q} = \{ \mathbf{q}_1,\dots,\mathbf{q}_m \} \gets \text{a new query}$
\State $\textbf{Output } \mathbf{f}_d^* \gets \text{a new array of } n \text{ zeros}$
\For{$i \gets 1 \text{ to }m$}
\State $\textrm{obtain } \textrm{NN}^{\textrm{SIM}}_k(\mathbf{q}_i), \textrm{NN}^{\textrm{ID}}_k(\mathbf{q}_i)  \text{ by $k$-NN search}$
\For{$j \gets 1 \text{ to } k$}
\State $\mathtt{col\_id} \gets \textrm{NN}^{\textrm{ID}}_k(\mathbf{q}_i)[j]$
\State $\mathtt{weight} \gets \textrm{NN}^{\textrm{SIM}}_k(\mathbf{q}_i)[j]$
\State $\mathtt{row\_ids} \gets \textrm{NN}^{\textrm{ID}}_L(\mathbf{x}_\mathtt{col\_id}) \text{ from sparsified } \mathcal{L}_\alpha^{-1}$
\State $\mathbf{f}_d^*[\mathtt{row\_ids}] \gets \mathbf{f}_d^*[\mathtt{row\_ids}] + \mathtt{weight} * \hat{\mathbf{c}}_{\mathtt{col\_id}}$
\EndFor
\EndFor
\State $\text{Aggregate scores in } \mathbf{f}_d^* \text{ to image level if needed}$
\end{algorithmic}
\end{algorithm}

\subsection{Online search}

Once the sparsified inverse matrix is obtained, the online search results can be calculated by using~\cref{alg:online}.
We denote $\textrm{NN}_k^{\textrm{SIM}}(\mathbf{q}_i)$ as the similarities between the query feature $\mathbf{q}_i$ and its nearest neighbors in $\chi$.
The online search thus becomes very fast because it only includes the $k$-NN search and linear combination.
Moreover, since we only need $k$ columns in $\mathcal{L}_\alpha^{-1}$ for each query, our approach can merely cost $\mathcal{O}(Lk)$ RAM during runtime.

\section{Experiments}

This section presents the experimental setup and investigates the computational efficiency as well as retrieval performance of our methods for image retrieval.
For the efficiency evaluation, we use a single core of Intel Xeon 2.80GHz CPU.

\subsection{Experimental setup}

\subsubsection{Datasets.}
We use the Oxford Buildings~\cite{philbin2007object} and Paris~\cite{philbin2008lost} datasets in our experiments.
The datasets are referred to as Oxford5k and Paris6k respectively in correspondence with the size of each dataset.
Another set of 100k random images from Flicker~\cite{philbin2008lost} are commonly used as distractors to enlarge the above datasets to Oxford105k and Paris106k.
We measure the online computational time on the 55 queries of the datasets for Iscen's method~\cite{iscen2017efficient} and our proposed method.
For evaluation, we adopt the standard mean average precision (mAP) as a performance measurement.

\subsubsection{Features.}
We use the 512 and 1,024 dimensional global R-MAC descriptors~\cite{tolias2015particular,gordo2016deep} provided by Iscen et al. for fair comparison.
We experiment on both global and regional features provided.
For the Oxford and Paris datasets, there are 21 regional features per image on average.  

\subsubsection{K-NN search.}
We conduct $k$-NN search by using the efficient FAISS toolkit~\footnote{\url{https://github.com/facebookresearch/faiss}}, containing a CPU version and a faster GPU version~\cite{johnson2017billion}, which allows us to deal with the larger Oxford105k and Paris106k datasets, especially for offline pre-computation.

\begin{figure}[t]
  \centering
  \begin{subfigure}[b]{\linewidth}
    \includegraphics[width=\linewidth]{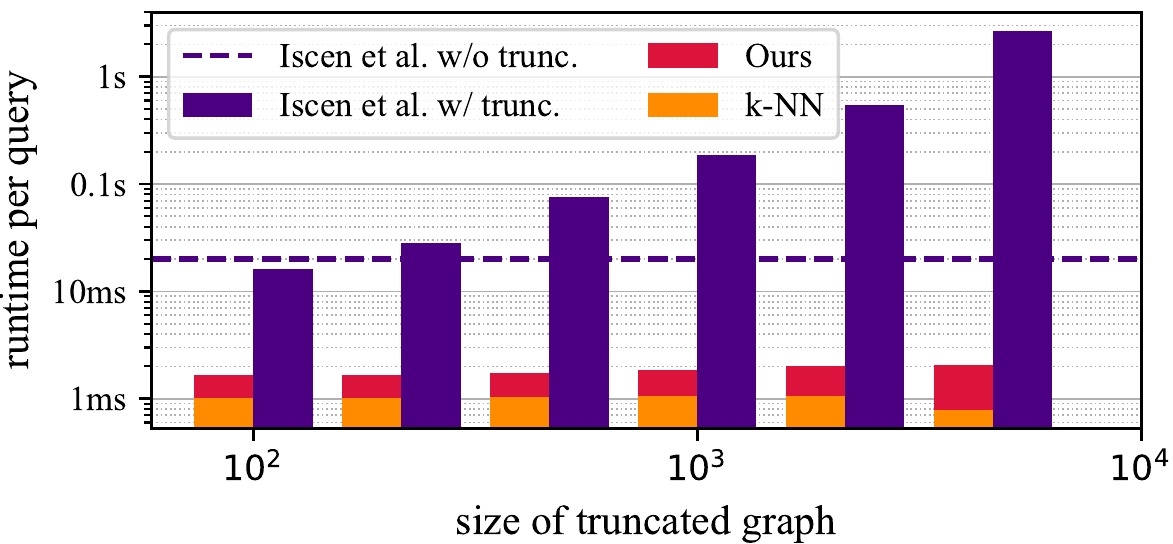}
    \caption{Oxford5k}
  \end{subfigure}
  \begin{subfigure}[b]{\linewidth}
    \includegraphics[width=\linewidth]{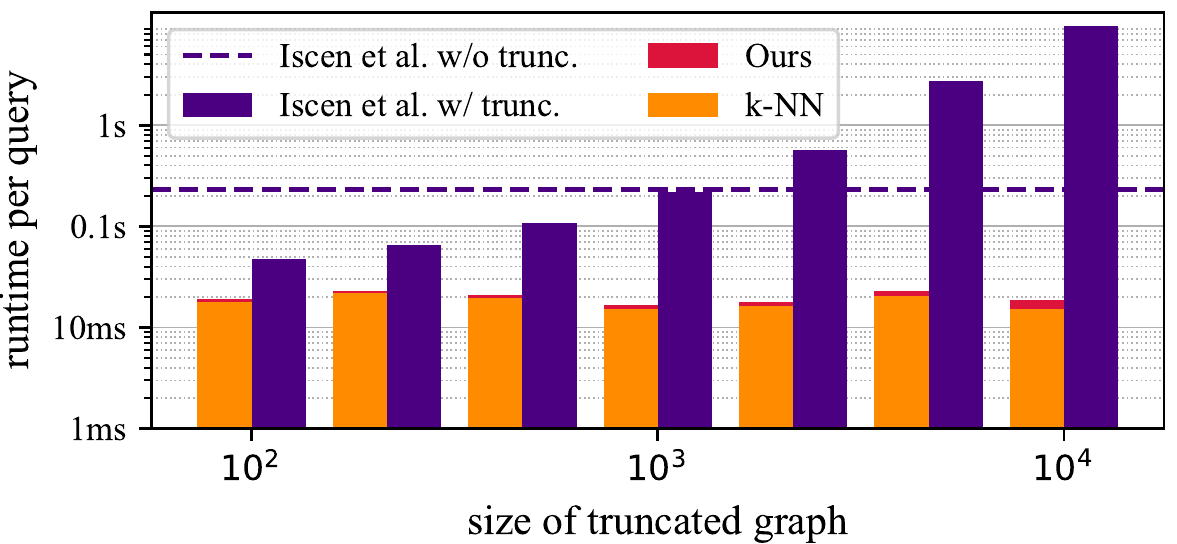}
    \caption{Oxford105k}
  \end{subfigure}
  \caption{Online search time per query measured for $k$-NN search, Iscen's method and our proposed method respectively, using global features.}
  \label{fig:time}
\vspace{-3mm}
\end{figure}

\begin{figure}[t]
  \centering
  \includegraphics[width=.8\linewidth]{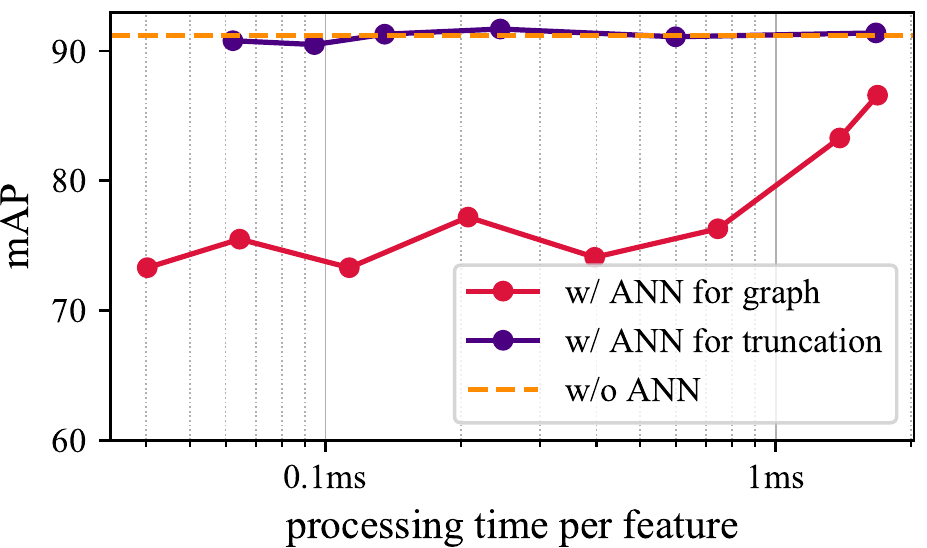}
  \caption{Speed/accuracy trade-off by using approximate nearest neighbor (ANN) search in truncation and graph construction.}
  \label{fig:offline}
\vspace{-3mm}
\end{figure}

\subsubsection{Implementation details.}
We use the same graph construction parameters as in the previous work~\cite{iscen2017efficient}.
In particular, the parameter $\alpha$ to build $\mathcal{L}_\alpha$ is set to $0.99$.
For global features, 50 nearest neighbors of each database element are used for graph construction, and the initial state vector contains the similarities between the query and its 10 nearest neighbors.
While for regional features, the corresponding numbers of nearest neighbors are set to 200, 200 respectively.
Through our experiments, deviating from these parameters consistently resulted in worse performance.

\subsection{Runtime computational efficiency}

We evaluate the computational efficiency on each query for $k$-NN search, Iscen's method and our proposed method.
Each method is run 10 times and the average computational time is used for comparison.
The results on Oxford5k and Oxford105k datasets are shown in~\cref{fig:time}.
Since most of the computation in our proposed method is already done offline on database side, we observe that our method can be remarkably fast during online search, close to the speed of $k$-NN search.
\cref{fig:time} shows that the average search time per query of our method with global features are $\sim$2ms and $\sim$10ms on Oxford5k and Oxford105k datasets respectively, and its extra computation over $k$-NN search is negligible.
In contrast, Iscen's method is rather time-consuming since it has a lot of runtime processes.
The search time per query for online diffusion without truncation is slower: $\sim$20ms on Oxford5k and $\sim$0.2s on Oxford105k.
When truncation is applied during the offline diffusion process, the overhead to construct the graph during runtime causes it to be slow. We also observed a decrease in efficiency as the size of truncated graph grows as shown in ~\cref{fig:time}.

\begin{figure*}[t!]
\centering
\begin{subfigure}[b]{0.24\linewidth}
  \includegraphics[width=\linewidth]{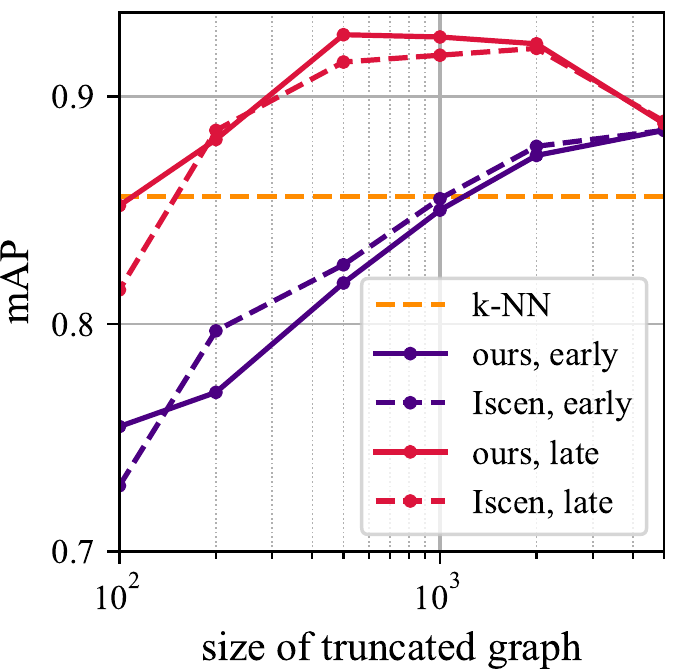}
  \caption{Oxford5k}
\end{subfigure}
\begin{subfigure}[b]{0.24\linewidth}
  \includegraphics[width=\linewidth]{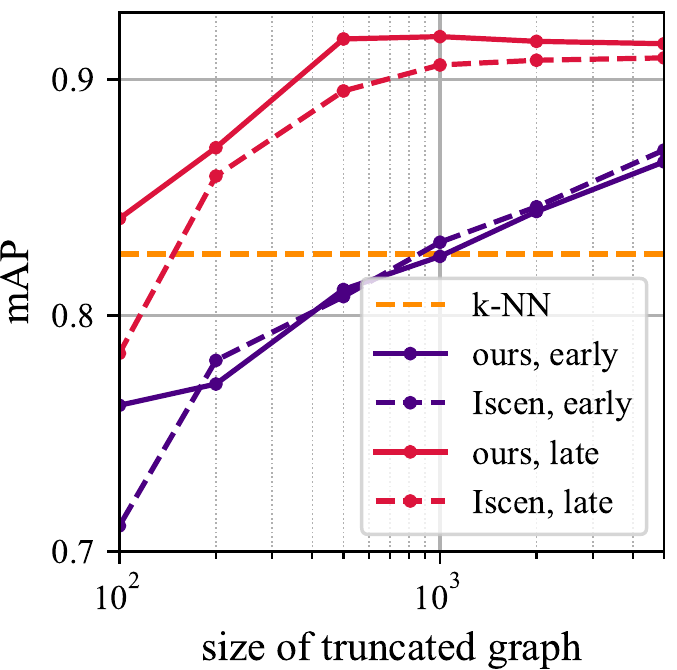}
  \caption{Oxford105k}
\end{subfigure}
\begin{subfigure}[b]{0.24\linewidth}
  \includegraphics[width=\linewidth]{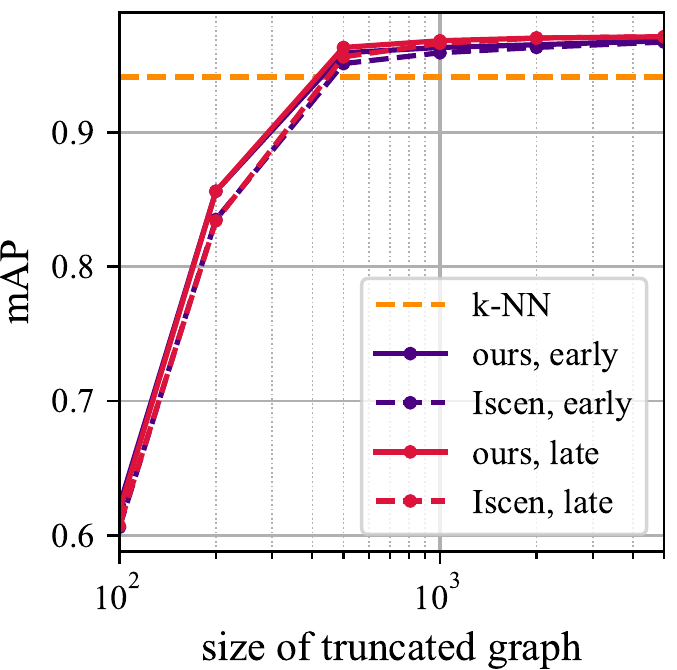}
  \caption{Paris6k}
\end{subfigure}
\begin{subfigure}[b]{0.24\linewidth}
  \includegraphics[width=\linewidth]{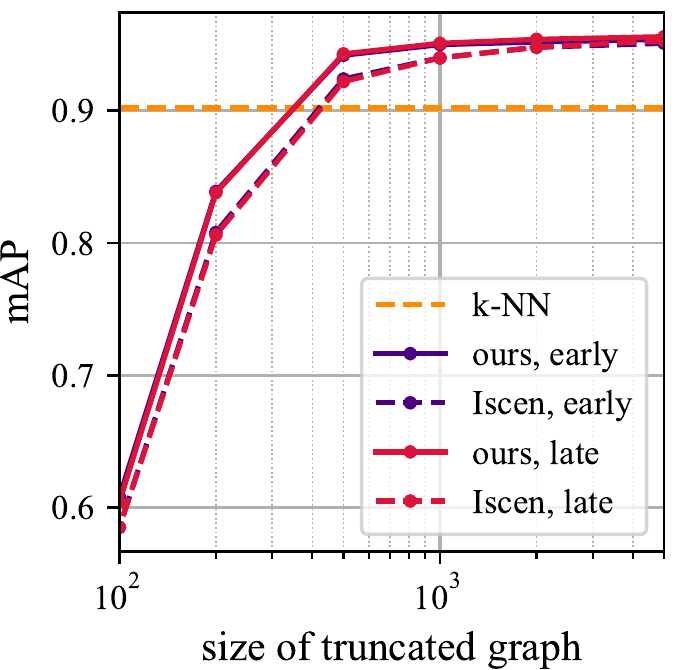}
  \caption{Paris106k}
\end{subfigure}
\caption{Retrieval performance (mAP) vs. the size of truncated graph $L$ using early truncation and late truncation.}
\label{fig:subgraph_norm}
\vspace{-2mm}
\end{figure*}

\begin{table*}[t]
\begin{center}
\begin{tabular}{@{}l l c c c c c c c@{}} \toprule
    & Method &  Feature & Global & Regional & Oxf5k  & Oxf105k  & Par6k & Par106k \\ \midrule
\multirow{8}*{\rotatebox[origin=c]{90}{w/o regional feature}} & k-NN search & \multirow{4}{*}{R-MAC (VGG)}
                                                      & \cmark &  & 79.5 & 72.1 & 84.5 & 77.1 \\
    & k-NN + AQE~\cite{chum2007total}               & & \cmark &  & 85.4 & 79.7 & 88.4 & 83.5 \\
    & Iscen's diffusion~\cite{iscen2017efficient}    & & \cmark &  & 85.7 & 82.7 & 94.1 & 92.5 \\
    & \textbf{Proposed diffusion}                   & & \cmark &  & \textbf{89.7} & \textbf{86.8} & \textbf{94.7} & \textbf{92.9} \\
 \cmidrule(lr){2-9}
    & k-NN search & \multirow{4}{*}{R-MAC (ResNet)}   & \cmark &  & 83.9 & 80.8 & 93.8 & 89.9 \\
    & k-NN + AQE~\cite{chum2007total}               & & \cmark &  & 89.6 & 88.3 & 95.3 & 92.7 \\
    & Iscen's diffusion~\cite{iscen2017efficient}    & & \cmark &  & 87.1 & 87.4 & 96.5 & 95.4 \\
    & \textbf{Proposed diffusion}                   & & \cmark &  & \textbf{92.6} & \textbf{91.8} & \textbf{97.1} & \textbf{95.6} \\
\midrule
    \multirow{10}{*}{\rotatebox[origin=c]{90}{w/ regional feature}} & R-match~\cite{razavian2016visual} & \multirow{5}{*}{R-MAC (VGG)} 
                                                      &        & \cmark & 81.5 & 76.5 & 86.1 & 79.9 \\
    & R-match + AQE~\cite{chum2007total}            & &        & \cmark & 83.6 & 78.6 & 87.0 & 81.0 \\
    & Iscen's diffusion~\cite{iscen2017efficient}    & & \cmark & \cmark & 93.2 & 90.3 & \textbf{96.5} & 92.6 \\
    & \textbf{Proposed diffusion}                   & &        & \cmark & 91.8 & 88.6 & 93.9 & 89.2 \\
    & \textbf{Proposed diffusion w/ late fusion}         & & \cmark & \cmark & \textbf{93.5} & \textbf{91.2} & 96.1 & \textbf{93.8} \\
 \cmidrule(lr){2-9}
    & R-match~\cite{razavian2016visual} & \multirow{5}{*}{R-MAC (ResNet)} 
                                                      &        & \cmark & 88.1 & 85.7 & 94.9 & 91.3 \\
    & R-match + AQE~\cite{chum2007total}            & &        & \cmark & 91.0 & 89.6 & 95.5 & 92.5 \\
    & Iscen's diffusion~\cite{iscen2017efficient}    & & \cmark & \cmark & 95.8 & 94.2 & 96.9 & 95.3\\
    & \textbf{Proposed diffusion}                   & &        & \cmark & 95.9 & 94.8 & 97.6 & 95.6 \\
    & \textbf{Proposed diffusion w/ late fusion}         & & \cmark & \cmark & \textbf{96.2} & \textbf{95.2} & \textbf{97.8} & \textbf{96.2} \\
\bottomrule
\end{tabular}
\end{center}
\caption{Performance comparison with the state of the art. We used R-MAC features extracted with VGG~\cite{radenovic2016cnn} and ResNet101~\cite{gordo2016deep}.}
\label{tab:results}
\vspace{-3mm}
\end{table*}

\subsection{Pre-computational efficiency}

Now that we have shown the online search of our proposed approach is very efficient, we investigate the offline computational cost.
The offline process mainly consists of two parts: the database vs. database $k$-NN search for truncation and random walk processes on each of the database element.
Since exhaustive $k$-NN search on a large scale dataset is time-consuming, we utilize approximate nearest neighbor (ANN) search.
Figure~\ref{fig:offline} shows the speed/accuracy trade-off when using ANN search for either truncation or graph construction.
We adopt IVFADC~\cite{jegou2011product} for ANN search using Faiss library~\cite{johnson2017billion}, where the codebook size of coarse quantizer $\sqrt{n}\approx316$, the number of subvectors $M=128$ is used and the number of clusters to scan is varied.
When compared to the exhaustive $k$-NN search, ANN is generally good at approximating the top results of the search but its ranking and scores can be out of order and imprecise.
This makes it applicable to truncation which only requires a coarse set of the top results.
Graph construction, however, would be negatively affected by even small differences in the scores of the search results.
We therefore use ANN search only for truncation and use exhaustive $k$-NN search for graph construction.

Since the $k$ in $k$-NN search for graph construction is small compared to truncation, it can be efficiently processed even without using approximate search, especially by using Faiss on a GPU.
As a result, the entire process of truncation and graph construction takes $\sim$1ms using a single GPU per image.
Diffusion processes take $\sim$6 minutes to process the whole Oxford105k dataset using global features and a truncation size of 5,000 (3.4 ms per image).
We measured using a single core of CPU, but these processes can be easily parallelized.
Compared to the offline processing in \cite{iscen2018fast} which takes a few hours, our method is much faster.

\subsection{Influence of subgraph normalization}

In~\cite{iscen2017efficient}, truncation enables diffusion on large scale datasets but is described to be detrimental to retrieval performance.
The authors claim that retrieval performance of diffusion grows as the percentage of the whole graph used grows, with a complete graph without any truncation having the best performance.
However, we argue that the retrieval performance is mainly influenced by the early truncation leading to subgraph normalization.
In order to avoid subgraph normalization, we first obtain the complete graph and apply late truncation (slicing) on the complete matrix $\mathcal{L}_\alpha$ in the process of diffusion.
For comparison, we implement Iscen's method with and without subgraph normalization.
We vary the truncation size $L$ to observe the influence of subgraph normalization on the retrieval performance.

The experimental results with global features for the Oxford and Paris datasets are presented in~\cref{fig:subgraph_norm}.
On the Oxford datasets, it is clear that the performance is significantly improved without subgraph normalization when the truncation size $L$ is small, but its effectiveness on the Paris datasets is smaller.
We observe that merely performing $k$-NN search on the Paris datasets already results in a high retrieval performance.
This could mean that the Paris datasets have standard-shaped manifolds conducive to comparison by Euclidean distance, so it limits the benefits gained from diffusion.

While previous works always encouraged using larger values of $L$ for performance gains, our method can achieve state-of-the-art performance with smaller $L$ values found through validation.
As a bonus, a smaller truncation size $L$ will lead to acceleration of the offline computation.

\subsection{Comparison to other methods}
\cref{tab:results} compares our method with other competitive methods that use global and regional features. Testing on all datasets using global features, we observe up to a 5\% increase in mAP performance compared to the previous state-of-the-art. Similarly, on diffusion with regional features, we achieve competitive or better performance.
We note that Iscen et al. used global features to guide the truncation in their regional diffusion.
For each query, they first apply $k$-NN search using the global features to obtain the closest images to that query.
Subsequently, these results are used as a bounded set to perform regional diffusion.
In contrast, we only use regional features in our regional diffusion for simplicity.
To exploit global features, we apply a simple late fusion by computing a weighted mean of scores from regional and global diffusion, setting the weight for regional diffusion to 0.75.
This further increases the performance (\textit{proposed diffusion w/ late fusion} in~\cref{tab:results}), leading to better performance than Iscen et al. on all datasets.

\section{Conclusion}

In this paper, we propose a novel efficient diffusion to achieve fast retrieval during runtime with significant improvement in retrieval performance.
We experimentally show that our approach has a similar efficiency to $k$-NN search, which is 10$\sim$ times faster than existing diffusion methods with global features.
Moreover, our method achieves state-of-the-art performance on Oxford and Paris datasets.
In conclusion, our method makes diffusion more practical for image retrieval on large-scale datasets, and has the potential to improve retrieval in other fields, such as text and video.

\section{Acknowledgment}

This study is supported by CREST (JPMJCR 1686).

\clearpage

\bibliographystyle{aaai}
\bibliography{refs}

\end{document}